\documentclass[conference]{IEEEtran}
\IEEEoverridecommandlockouts
\usepackage{cite}
\usepackage{amsmath,amssymb,amsfonts}
\usepackage{algorithmic}
\usepackage{graphicx}
\usepackage{textcomp}
\usepackage{xcolor}
\usepackage{multirow}
\usepackage{url}
\def\BibTeX{{\rm B\kern-.05em{\sc i\kern-.025em b}\kern-.08em
    T\kern-.1667em\lower.7ex\hbox{E}\kern-.125emX}}
\begin{document}

\title{Is Micro-expression Ethnic Leaning?}



\author{
    \IEEEauthorblockN{Huai-Qian Khor\IEEEauthorrefmark{3},
                      Yante Li\IEEEauthorrefmark{3},
                      Xingxun Jiang\IEEEauthorrefmark{2}, Guoying Zhao\IEEEauthorrefmark{3}\IEEEauthorrefmark{1}}
    \IEEEauthorblockA{\IEEEauthorrefmark{3}Center for Machine Vision and Signal Analysis, University of Oulu, Finland \\
                      Email: huai.khor@oulu.fi, yante.li@oulu.fi}
    \IEEEauthorblockA{\IEEEauthorrefmark{2}School of Biological Sciences and Medical Engineering, Southeast University \\
                      Email: jiangxingxun@seu.edu.cn}

$^\star$Corresponding Author, E-mail: guoying.zhao@oulu.fi
}

\maketitle

\begin{abstract}
How much does ethnicity play its part in emotional expression? Emotional expression and micro-expression research probe into understanding human psychological responses to emotional stimuli, thereby revealing substantial hidden yet authentic emotions that can be useful in the event of diagnosis and interviews. While increased attention had been provided to micro-expression analysis, the studies were done under Ekman's assumption of emotion universality, where emotional expressions are identical across cultures and social contexts. Our computational study uncovers some of the influences of ethnic background in expression analysis, leading to an argument that the emotional universality hypothesis is an overgeneralization from the perspective of manual psychological analysis. In this research, we propose to investigate the level of influence of ethnicity in a simulated micro-expression scenario. We construct a cross-cultural micro-expression database and algorithmically annotate the ethnic labels to facilitate the investigation. With the ethnically annotated dataset, we perform a prima facie study to compare mono-ethnicity and stereo-ethnicity in a controlled environment, which uncovers a certain influence of ethnic bias via an experimental way. Building on this finding, we propose a framework that integrates ethnic context into the emotional feature learning process, yielding an ethnically aware framework that recognises ethnicity differences in micro-expression recognition. For improved understanding, qualitative analyses have been done to solidify the preliminary investigation into this new realm of research. Code is publicly available at \url{ https://github.com/IcedDoggie/ICMEW2025_EthnicMER}
\end{abstract}




\vspace{-1em}

\section{Introduction}
Micro-expressions (MEs) are spontaneous, subtle and rapid (1/25 to 1/3 seconds) \cite{oh2018survey} facial movements that reveal true underlying and suppressed emotions. They are solicited under high-stakes environments \cite{ekman2009lie, ekman1971constants} such as psychological diagnosis, interviews and interrogations. While MEs research had gained increasing attention in the field of affective computing and behavioral analysis, conventional research had emphasised the spotting and recognition tasks \cite{li2022deep} under the assumption of emotion universality, as proposed by Ekman \cite{ekman1987universals}. Meanwhile, opposed to the hypothesis of emotion universality, ethnic influence could play a part in influencing the analysis, but it has yet to be given much attention in the MEs research domain \cite{see2024megc2024}.  


The concept of emotional universality remains a contentious topic within facial expression (including micro-expression) analysis. Ekman et al. \cite{ekman1987universals} posit that emotional expressions are consistent across cultures and ethnic groups; in other words, emotion universality exists. While the universality hypothesis has provided a foundational framework for emotion research, emerging studies have begun to look into the ethnicity element, particularly in the context of macro-expressions \cite{chen2024cultural}, to study the potential bias or uncover additional insights in emotion expression across cultures. On another perspective, the generalizability of current facial expression analysis that happens across diverse and vibrant populations is also questionable, as perception towards a stimulus or social context can vary across individuals \cite{cowen2021sixteen}. Despite that, the intersection of facial expressions and cultural influences has still barely received scholarly attention \cite{anderson2018emotion} \cite{tracy2008spontaneous} as there is a limited ethnically-targeted facial analysis data.

Despite the growing awareness of ethnic and cultural bias in affective computing, there remains a significant gap in the literature when it comes to understanding how ethnic factors influence micro-expression analysis. Given the subtle and transient nature of MEs, it is plausible that such expressions are even more susceptible to cultural variability and perceptual bias, both in human observers and algorithmic systems. The earliest cross-cultural ME analysis \cite{see2024megc2024} examined the impact of cultural variation on micro-expression spotting. Their findings suggest that, under the assumption of a consistent data shift, cultural factors do influence expression analysis. Beyond this, the role of cultural context in micro-expression recognition remains largely unexplored and lacks formal investigation.

In this paper, we explore the impact of cross-cultural factors on micro-expression analysis within a pseudo-simulated cross-database environment. Our contributions are as follows:
\begin{itemize}
    \item Establishing a combined dataset using CASME II and SAMM, enriched with ethnic annotations obtained semi-autonomously, subsequently enabling a systematic exploration of the relationship between ethnic and micro-expression characteristics. 
    \item Investigating cross-cultural influence in micro-expression recognition and exploring the correlation between ethnic background and micro-expression characteristics.
    \item Proposing a framework that incorporates ethnicity-based features alongside facial motion features to study the efficacy of ethnic integration in feature study.
\end{itemize}

\section{Related Work}

In a psychological study, Cowen et al. \cite{cowen2021sixteen} analysed data from 144 countries to examine the relationship between social contexts (such as fireworks, weddings, and sporting events), facial expressions, and global regions. Their findings demonstrated that 70\% of context-expression associations were consistent across 12 world regions, supporting the notion of universal facial responses to specific social stimuli. However, the study also revealed geographical variations in facial expressions, with neighbouring regions often exhibiting more similarity than those further apart. Conversely, Chen et al. \cite{chen2024cultural} examined the differences in the six basic emotional expressions—happiness, surprise, fear, disgust, anger, and sadness—between East Asian and Western European populations. Their findings revealed that while both cultures exhibited similar patterns for threat-related emotions like anger and fear, they differed in the expression of low-threat emotions such as happiness and sadness, particularly in terms of timing and facial Action Unit (AU) activation. These works serve as the basis of investigation for us to look into cross-cultural aspects in micro-expression.

In computational research, cross-cultural facial analysis focuses on narrowing the cultural gap in feature learning by either merging datasets or conducting hold-out database evaluations. Notably, in the field of micro-expressions, MEGC 2023 \cite{see2024megc2024} hosted a challenge aimed at detecting micro-expressions within a cross-cultural context. The challenge involved combining the SAMM dataset, which features multicultural subjects, with CAS(ME)$^3$, which consists exclusively of Asian subjects, to evaluate the robustness of micro-expression spotting across cultures. In another domain, Xu et al. \cite{xu2023humor} investigated cross-cultural humour detection by training on German data and testing on English. Their study aimed to bridge the linguistic and cultural gap in modelling humour, specifically within the context of football interview videos. While these two works serve as the early or landmarked works, they lack an in-depth study of how ethnicity can affect the eventual recognition performance specifically. 

As previously noted, the correlation between facial expressions and culture has received limited attention, especially in the context of micro-expressions. While traditional research has largely focused on improving performance metrics, other critical aspects, such as cultural influence, have been largely overlooked. This narrow focus introduces potential biases that may hinder the broader applicability and reliability of future real-world deployments.

In this work, we aim to construct an ethnicity-annotated joint micro-expression database to investigate the relationship between ethnic identity and micro-expression emotion analysis. Building upon the foundation laid by \cite{see2024megc2024}, we enhance the exploration of cross-cultural factors by incorporating explicit ethnic annotations. Additionally, we conduct a prima facie study to further substantiate the presence of ethnic influence in micro-expression analysis. Ultimately, we integrate ethnic labels into the feature learning process, enabling the model to leverage both emotional and ethnic contexts in micro-expression prediction.


\section{Methodology}
The methodology dives into the details of three aspects: (1) how the ethnic-annotated joint dataset is constructed, (2) the prima facie design to validate the existence of ethnic influence in micro-expression analysis and (3) the proposed simplistic framework to incorporate ethnicity context into feature learning. 

\subsection{Dataset Construction \& Automated Annotation}
\begin{figure}[!ht]
    \centering
    \includegraphics[scale=0.5]{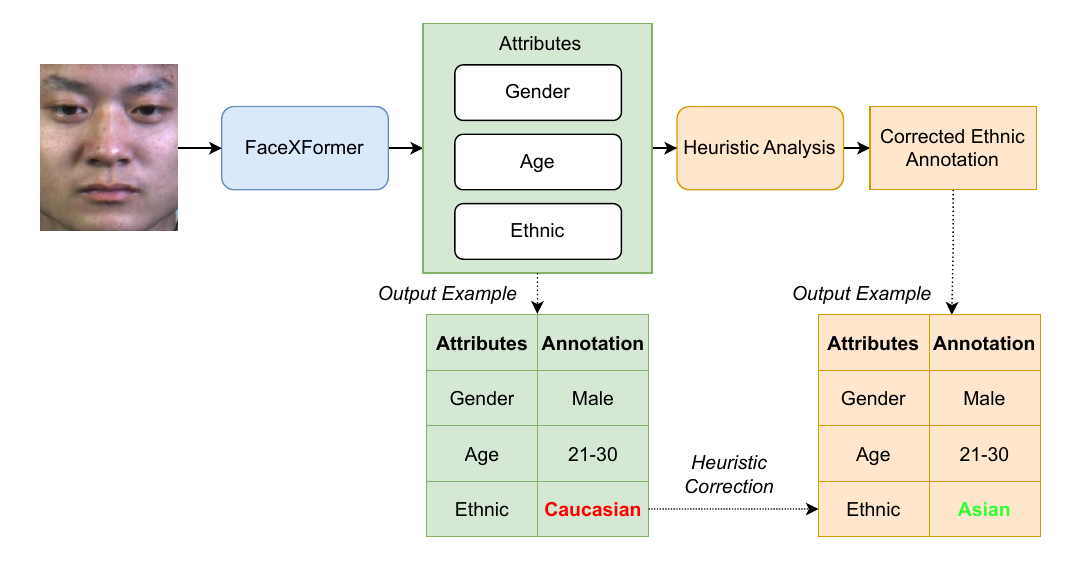}
    \caption{The process of annotation, where FaceXFormer extracts gender, age and ethnicity of the subjects. Then, a heuristic screening is performed to correct some of the labels. The output examples of the annotation and correction are demonstrated via the dotted lines in the diagram.}
    \label{fig:annotation_process}
\end{figure}

To construct a simulated dataset, we employ FaceXFormer \cite{narayan2024facexformer} to annotate the ethnicity of micro-expression samples into five categories: Caucasian (White), African (Black), Asian, Indian, and Others. In detail, we first utilise FaceXFormer to annotate the attributes such as gender, age and ethnicity. Following that, we perform a heuristic-driven screening to correct some annotations across all attributes, as the algorithm may predict the wrong ethnicity in contrast to human eyes. The process is as demonstrated in the Figure \ref{fig:annotation_process}. While the heuristic data correction is performed on all attributes (gender, age and ethnicity), the gender and age of subjects are exempted from the study as our scope of research is on ethnicity.

Altogether, the combined CASME2 and SAMM datasets include 54 unique subjects. The original ethnic labels, along with their corresponding mapped labels (mapping is done for data distribution balancing), are presented in Table \ref{tab:subject_samples}.

\begin{table}[!ht]
    \centering
    \caption{Number of subject samples per ethnicity.}
    \begin{tabular}{|c|c|c|}
        \hline    
         Ethnic Labels & Mapped Labels & Number of Subject Samples\\
         \hline
         Caucasian(White) & Non-Asian & 15 \\
         African(Black) & Non-Asian & 2 \\
         Asian & Asian & 31 \\
         Indian & Asian & 2 \\
         Others & Asian & 4 \\
         \hline
    \end{tabular}
    \label{tab:subject_samples}
\end{table}

To balance the data distribution, we consolidate these labels into two broad groups: Asian (including both Asian and Indian) and non-Asian (including Caucasian and African). This grouping is based on geographical considerations, where ethnicities originating from Asia are categorised as Asian, and those from outside Asia are classified as non-Asian. The case of Asian with non-Asian origin, and vice versa, will be exempted from this study as it has no data support for this use case. Table \ref{tab:race_label} shows the video samples for each unique ethnicity and the concise ethnic labels.

\begin{table}[!ht]
    \centering
    \caption{Number of video samples per mapped ethnicity labels. ``Others" is classified as Asian here via manual heuristic assessment.}
    \begin{tabular}{|c|c|c|}
        \hline    
         Ethnic Labels & Mapped Labels & Number of Video Samples\\
         \hline
         Caucasian(White) & Non-Asian & 88 \\
         African(Black) & Non-Asian & 4 \\
         Asian & Asian & 183 \\
         Indian & Asian & 5 \\
         Others & Asian & 11 \\
         \hline
    \end{tabular}
    \label{tab:race_label}
\end{table}

To integrate the two datasets and streamline emotional classification, we remap the emotion labels into three categories: Positive, Negative, and Surprise. The "Others" category is excluded to reduce ambiguity in emotional interpretation in our study. The data distribution is illustrated in Table \ref{tab:emotion_label}. While it demonstrates a certain degree of imbalance for the mapped labels, and possibly a low data amount for a statistical study, the simplification of the emotion and ethnic labels that we have made enables feature learning to be performed to study the influence of ethnic context in emotion analysis.

\begin{table}[!ht]
    \centering
    \caption{Number of video samples per mapped emotion labels. ``Others" is omitted as the emotional context is ambiguous. }
    \scalebox{0.95}{
    \begin{tabular}{|c|c|c|}
        \hline    
         Emotion Labels & Mapped Labels & Number of Video Samples\\
         \hline
         Happiness & Positive & 58\\
         \hline
         Anger, Contempt, Disgust  & \multirow{2}{*}{Negative} & \multirow{2}{*}{192} \\
         Fear, Repression, Sadness & & \\
         \hline 
         
         Surprise & Surprise & 40 \\
         \hline
         
         Others & - & - \\
         \hline
    \end{tabular}}
    \label{tab:emotion_label}
\end{table}

\subsection{Prima Facie Design}
To examine the presence of the issue, we conduct a prima facie study to compare the mono-ethnic and mixed-ethnic groups. To isolate the cultural factor and demonstrate the validity of the problem, we aim to minimise the potential impact of domain shift. Rather than evaluating individual databases separately, we design the study within a joint-database setting, applying ethnic-driven and cross-database evaluation (CDE) sampling strategies. This approach introduces stochasticity in domain variation while ensuring that the primary manipulated variable remains the subject’s ethnicity.

For this purpose, we have selected 16 subjects across three scenarios to maintain ethnic balance in the analysis:

\begin{itemize}
    \item Asian only. 16 Asian subjects are sampled from the combined CASME II–SAMM database.
    \item Non-asian only. 16 non-Asian subjects are sampled from the same joint database.
    \item Mixed Ethnicity. A balanced group of 8 Asian and 8 non-Asian subjects is selected from the joint database.
\end{itemize}



\subsection{Feature Extraction \& Fusion}
\begin{figure*}[!ht]
    \centering
    \includegraphics[scale=0.85]{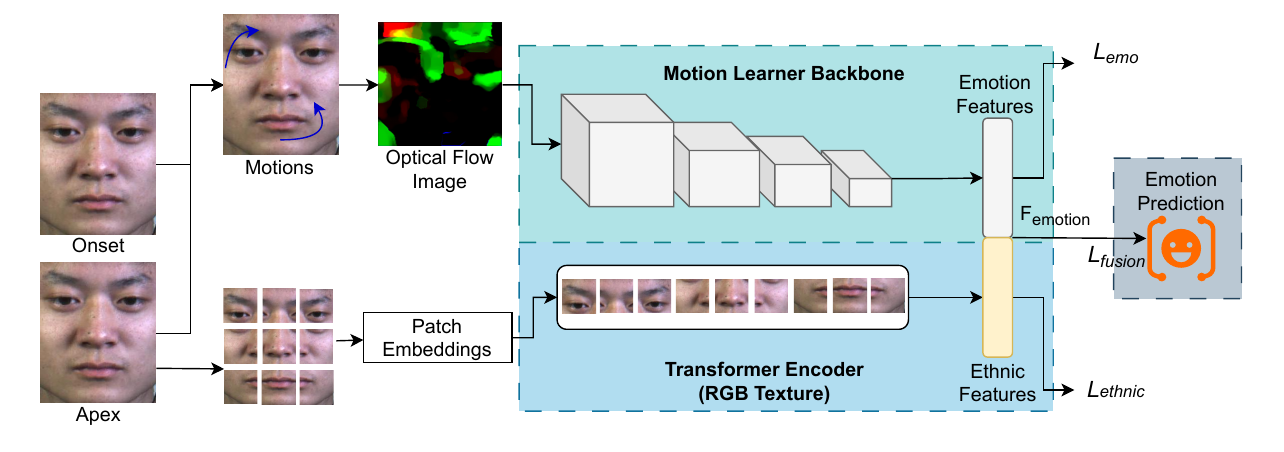}
    \caption{The backbone model is fine-tuned separately with different labels. The ethnic-driven backbone is fine-tuned with ethnic labels (Asian/Non-asian), whereas the emotion-driven backbone is fine-tuned with emotion labels (Positive/Negative/Surprise).}
  
    \label{fig:backbone_model}
\end{figure*}
For feature extraction, we design two perspectives on fusing the features. (1) Motion-based features to correlate the motions with expressions and a patch-based transformer to analyse the face texture to correlate with the ethnic context. (2) Motion-based features to correlate the motions with expressions and motion-based features that correlate the motions with the specified ethnic context.

Respectively to (1), the proposed approach aims to incorporate features learnt from emotion and ethnic features in two backbones with the motion input. The features are then fused to provide a ethnic context for micro-expression emotion prediction. Whereas for (2), the framework extracts patch-based embeddings from a RGB frame to fine-tune a vision transformer. This setup aims to investigate the correlation between RGB features and ethnicity, in comparison to motion features and ethnicity, based on the rationale that human perception is more attuned to identifying ethnicity through RGB visual cues.

As shown in Figure \ref{fig:backbone_model}, motion between the onset and apex frames is computed and summarised into an optical flow image. This image is then processed by the two separate backbones without weight-sharing to learn features specific to emotion and ethnic categories, respectively. The decision to forgo weight sharing is intentional, as it allows the model to decouple the learned representations for emotion and ethnicity. This separation enables the extraction of distinct feature perspectives and facilitates the explicit learning of ethnic context, which is subsequently integrated into the final classification as: $MergedFeatures = \{f_{Emotion}, f_{Ethnicity}\}$. Each of the emotions, ethnicities, and merged contexts is learned through its own fully connected layer, and the overall training objective is defined by the categorical cross-entropy loss function described in Equation \ref{eq:loss}. The emotion, ethnicity, and merged features are separately learnt in three distinct categorical cross-entropy (CCE) loss functions as shown in Equations \ref{eq:emoteloss}, \ref{eq:raceloss} and \ref{eq:mergedloss}, which is then combined in Equation \ref{eq:loss}.

\begin{equation}
    L_{emo} = CCE(Emote, \hat{Emote})
    \label{eq:emoteloss}
\end{equation}
\vspace{-1em}
\begin{equation}
    L_{ethnic} = CCE(Ethnicity, \hat{Ethnicity})
    \label{eq:raceloss}
\end{equation}
\vspace{-1em}
\begin{equation}
    L_{fusion} = CCE(Emote, Merged Features)
    \label{eq:mergedloss}
\end{equation}
\vspace{-1em}
\begin{equation}
    Loss = L_{emo} + L_{ethnic} + L_{fusion}
    \label{eq:loss}
\end{equation}

\textbf{Model Inputs}.
The input to the model consists of motion changes between the onset and apex frames. In our proposed feature extraction framework, these changes are represented as an optical flow image composed of three channels: the $x$ and $y$ displacements of motion vectors, and a third channel capturing optical strain features, which represent the second derivative of the motion vectors.

Optical flow is the first derivative of motion displacement, signified as $u$ for horizontal displacement and $v$ for vertical displacement. The computation is formalised in Equation \ref{eq:optical_flow}. 
\begin{equation}
	\vec{v} = [p = \frac{dx}{dt}, q = \frac{dy}{dt}] ^T
    \label{eq:optical_flow}
\end{equation}
where $dx$ and $dy$ represent the estimated changes in pixels along the $x$ and $y$ dimension respectively while d\textit{t} represent the change in time,

Optical strain \cite{shreve2009towards} is the derivative of optical flow, or the second-order derivatives of horizontal and vertical displacements. It is formalised in Equation \ref{eq:strain_short}.
\begin{equation}
	\epsilon = \frac{1}{2}[\nabla u + (\nabla u)^T]
    \label{eq:strain_short}
\end{equation}
where it calculates the second-order derivatives of horizontal and vertical displacements.

Given their representative capability in tiny motions, we utilise both optical flow\cite{sun2010secrets} and optical strain\cite{shreve2009towards} to compute the motion changes from onset to apex frames. To combine their representativeness, we group them as a feature cube known as an optical flow image (optical flow + optical strain) as formalised in Equation \ref{eq:optical_flow_image}.
\begin{equation}
    I_{OF} = ({f_x, f_y, \epsilon})
    \label{eq:optical_flow_image}
\end{equation}
where $I_{OF}$ is the optical flow image, $f_x$ and $f_y$ are the flow motion along the horizontal and vertical directions, respectively, and $\epsilon$ is the optical strain as computed in Equation \ref{eq:strain_short}.

\textbf{Feature Fusion}. Once the emotion and ethnic features have been learnt, the top-level features are merged, thereby concatenating both the emotion and ethnic context. The merged features are then passed to a separate fully connected layer, FC, to learn the dual-context features. To formalise, the process of fusion is demonstrated in Equation \ref{eq:fusion}.
\begin{equation}
    F_{emotion} = FC(concat(f_{emotion}, f_{ethnicity}))
    \label{eq:fusion}
\end{equation}

\section{Experiments}
The experiments are separated into a few sections: (1) A prima facie to study the existence of the problem, (2) benchmarking with several backbones with the CASME2-SAMM ethnically annotated dataset. 

For the benchmarking, experiments are conducted using a leave-one-subject-out (LOSO) protocol. To evaluate the combined CASME2-SAMM dataset, a composite database evaluation (CDE) approach is employed. The models are fine-tuned with 15 epochs with a learning rate of 0.001, Adam optimizer and ExponentialLR decay are utilised for the optimization process. The performance of each model is measured in macro-F1 score for a class-balanced evaluation.

\subsection{Prima Facie Results}
To identify the problem, we have created a subset of data from the CASME2-SAMM joint database, sampling 16 subjects for each test case. The reason for choosing 16 is that this is the maximum number of non-asian subjects. Using 16 can ensure that as many samples are available for the feature learning process, thereby limiting the effect of overfitting.

In terms of the expression classes, we have simplified the expressions into negative and non-negative. This is motivated to reduce the class imbalance influence, thereby restricting class imbalance issues from influencing the prima facie classification performance. Eventually, this prima facie study has 16 subjects with relatively balanced negative/non-negative labels.

The prima facie analysis is completed with ResNet-18 off-the-shelf features, optical flow image input, and a random forest classifier under the LOSO protocol. The motivation being is to mitigate the small sample size in this prima facie study; simplifying this training can reduce the risk of overfitting compared to fine-tuning an end-to-end deep model. This study excludes the ethnic context to scope down the analysis in a motion-features-only environment, assuming that ethnicity is barely influential in the emotion analysis.

The results of the prima facie study are presented in Table \ref{tab:prima_facie}. While class imbalance issues do exist in this prima facie study, the data class distribution is almost consistent for each of the experiments shown in the table. Based on the Table \ref{tab:prima_facie}, performance remains consistent across mono-ethnic groups, indicating minimal ethnic shift within the same ethnic category. However, a significant performance drop is observed in the mixed-ethnicity scenario, where we can observe that smaller emotion classes, such as Positive and Surprise are 0 in performance scores, highlighting the impact of ethnic variation on expression analysis.

\begin{table}[!ht]
    \centering
    \caption{This table validates the ethnic influence in expression analysis in a toy experiment. }
    \scalebox{0.9}{
    \begin{tabular}{|c|c|c|c|c|c|}
        \hline
         Train Data & Test Data & Protocol & Negative & Non-negative & Average \\
         \hline
         Asian & Asian & CDE & 0.4330 & 0.4762 & 0.4546 \\
         Non-Asian & Non-Asian & CDE & 0.6929 & 0.2642 & 0.4785 \\
         Mixed & Mixed & CDE & 0.8144 & 0.0606 & 0.4375 \\


         \hline
    \end{tabular}
    }

    \label{tab:prima_facie}
\end{table}

\subsection{Benchmarking}
\begin{table*}[!ht]
    \centering
    \caption{The benchmarking table that outlines the performance of several backbone models, consists of the presence and absence of ethnic context as an additional layer of features.}    
    \begin{tabular}{|c|c|c|c|c|c|c|c|c|}
        \hline
        Method & Motion Context & Ethnic Context & Ethnicity Representation & Negative & Positive & Surprise & Average MF1 \\
        \hline
        ResNet-18 & \checkmark & X & N/A & 0.8142 & 0.5225 & 0.5263 & 0.6210 \\
        ResNet-18 & \checkmark & \checkmark & Optical Flow & 0.8520 & 0.5321 & 0.6076 & 0.6639 \\
        ResNet-18 & \checkmark & \checkmark & RGB Texture & 0.7798 & 0.2597 & 0.2985 & 0.4460 \\

        
        ResNet-18(M), TinyViT(T) & \checkmark & \checkmark & RGB Texture & 0.8405 & 0.5586 & 0.6486 & 0.6826 \\        

        \hline
    \end{tabular}

    \label{tab:benchmark}
\end{table*}
In feature extraction, we benchmark the CASME2-SAMM merged database with the LOSO protocol in convolutional-based and transformer-based backbones. For clarity purposes, we outline the motion context, ethnic context and how the ethnicity representation is learnt (whether we correlate optical flow motions or RGB Face Texture of Apex Frame with ethnic labels) in Table \ref{tab:benchmark}. Motion context signifies that the optical flow image is utilised, which is true for all cases; ethnic context signifies whether ethnicity is considered in emotion classification; ethnicity representation signifies the type of input features we use to fine-tune the model with the ethnic labels.

Based on the Table \ref{tab:benchmark}, the vanilla motion feature-based ResNet-18 achieves a significant score of 0.6210 average MF1 in a three-class micro-expression emotion classification as a baseline. Subsequently, the addition of ethnic context with dual ResNet-18 with late fusion increases the average MF1 by 4\%, indicating that the ethnic context correlating with motions provides an additional layer of features on top of the emotion-only method. As per the study of ethnic context via RGB visual cues, we have discovered that using ResNet-18 fails horribly in average MF1, indicating that RGB visual cues barely guarantee the ethnic context to be learnt. Therefore, we have tested with a transformer-based method that learns from a patch-based embedding of RGB faces, concatenating the transformer features with motion-learned ResNet-18 features. To our surprise, we observe that the average MF1 score has slightly improved compared to learning the ethnic context merely based on the correlation with motion features alone. This demonstrates that diving into a more locally based context improves the machine's understanding of RGB visual cues, thus correlating better with the ethnic context. 

\subsection{Correlation of Ethnic Context and Expression Activations}
In qualitative analysis, we investigate the correlation between the emotion's activation maps and the ethnic context. Across three emotion classes, we only study the positive and surprise classes because these emotion classes are monotonic, meanwhile, we opt out of studying the negative class because it is a built-up of multiple emotion classes such as disgust, fear and anger, as such, it will result in ambiguity in the eventual activation maps.

\begin{figure}[!ht]
\scalebox{0.5}{
\begin{tabular}{ccc}
  \includegraphics[width=0.3\textwidth]{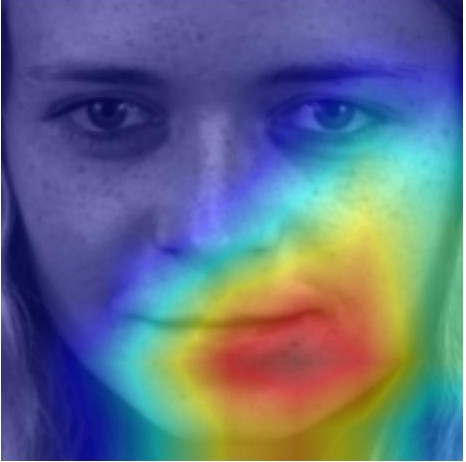} &
  \includegraphics[width=0.3\textwidth]{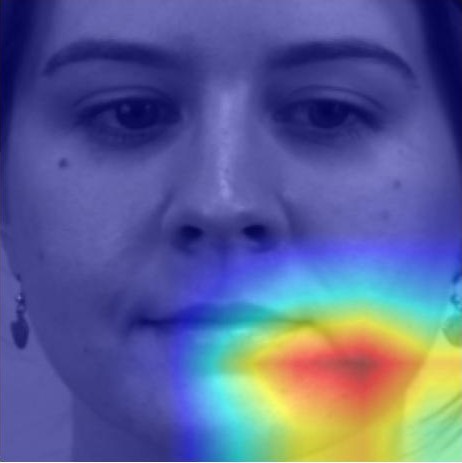} &
  \includegraphics[width=0.3\textwidth]{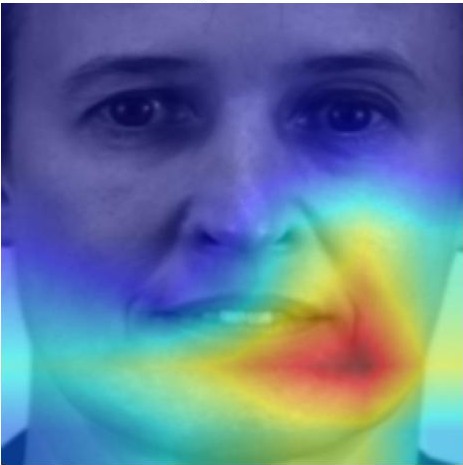} \\
  \includegraphics[width=0.3\textwidth]{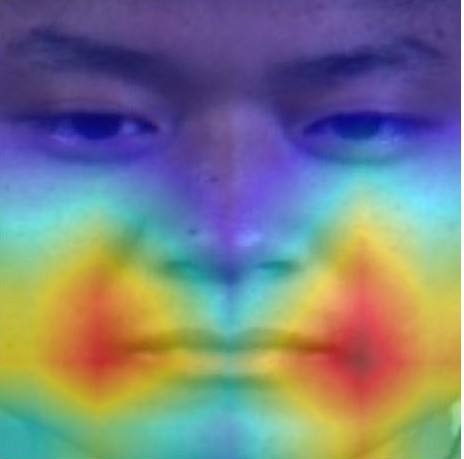} &
  \includegraphics[width=0.3\textwidth]{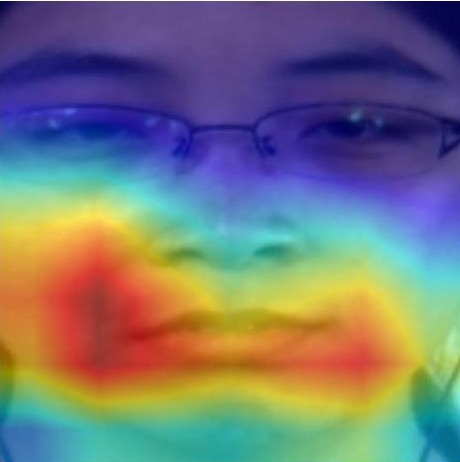} &
  \includegraphics[width=0.3\textwidth]{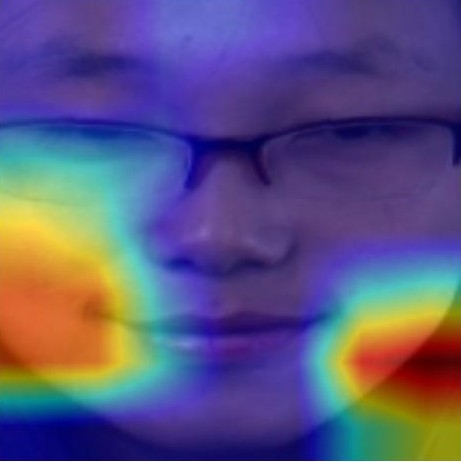} \\
\end{tabular}
}
\caption{The illustration of activation maps for positive classes. The upper row shows the samples belonging to non-asian, whereas the lower row represents asian samples. The corresponding action units for the positive (happiness) class consist of AU6 (Cheek Raiser) and AU12 (Lip Corner Puller).}
\label{fig:positive_gradcam}
\end{figure}

Based on Figure \ref{fig:positive_gradcam}, we can observe that for non-Asians, AU12 (Lip Corner Puller) is more active on one side of the face, and the activated region appears smaller despite it encompassing both AU6 (Cheek Raiser) and AU12 (Lip Corner Puller). Meanwhile, for Asians, we can observe that the activated regions tend to be slightly larger, similarly encompassing both AUs. One interesting difference is that Asians demonstrate activated maps on both sides of the face instead of one, as compared to non-Asians; this shows that there is emotional expression uniqueness across cultures, which defies the theory of emotion universality proposed by Ekman.

\begin{figure}[!ht]
\scalebox{0.5}{
\begin{tabular}{ccc}
  \includegraphics[width=0.3\textwidth]{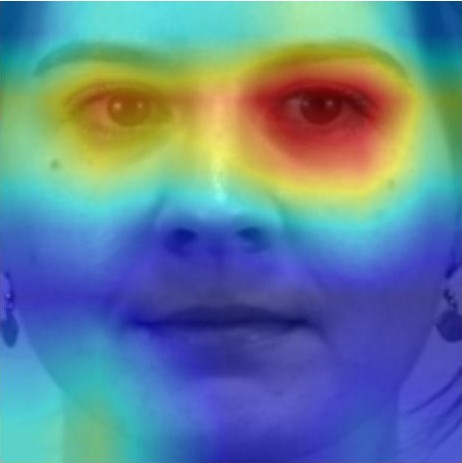} &
  \includegraphics[width=0.3\textwidth]{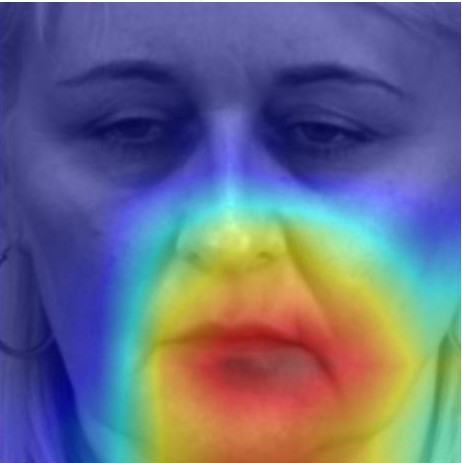} &
  \includegraphics[width=0.3\textwidth]{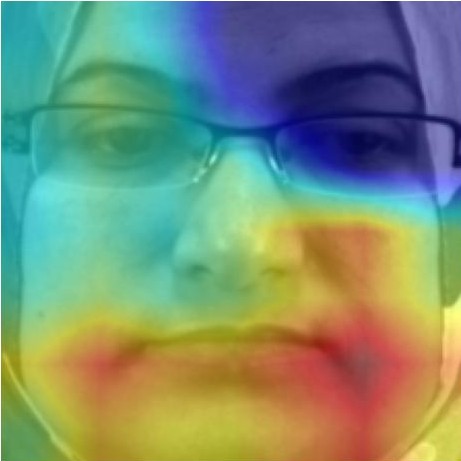} \\
  \includegraphics[width=0.3\textwidth]{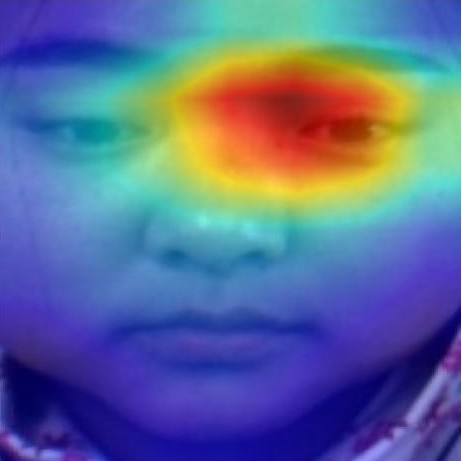} &
  \includegraphics[width=0.3\textwidth]{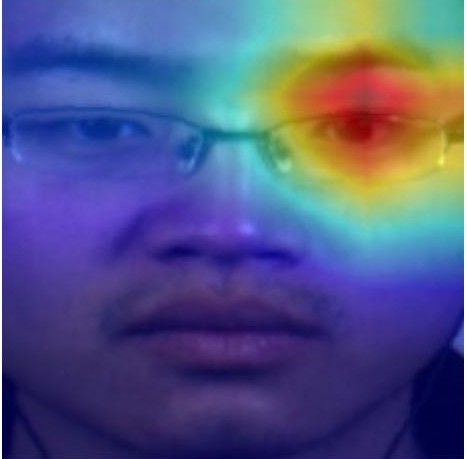} &
  \includegraphics[width=0.3\textwidth]{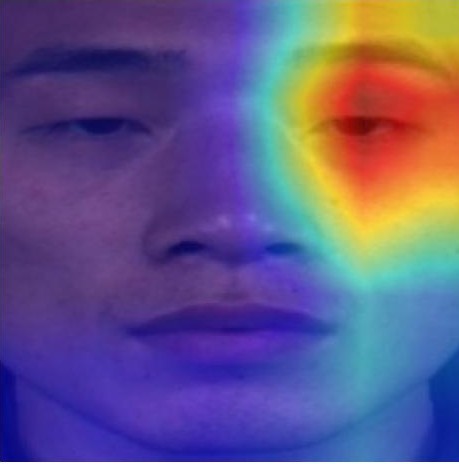} \\
\end{tabular}
}
\caption{The illustration of activation maps for surprise classes. The upper row shows the samples belonging to non-asian, whereas the lower row represents asian samples. The corresponding action units for the surprise class consist of AU1 (Inner Brow Raiser), AU2 (Outer Brow Raiser), AU5 (Upper Lid Raiser) and AU26 (Jaw Drop).}
\label{fig:surprise_gradcam}
\end{figure}

On the other side, referring to Figure \ref{fig:surprise_gradcam} for the surprise emotion, for non-Asians, we can observe that AU26 (Jaw Drop) is comparatively more active as compared to upper-faced AUs, signifying that non-Asians express surprise more commonly in a lower-faced region. Meanwhile, for Asians, upper-faced AUs such as AU1 (Inner Brow Raiser), AU2 (Outer Brow Raiser) and AU5 (Upper Lid Raiser) are more active compared to lower-faced AUs such as AU26 (Jaw Drop). Comparing across cultures, it shows that different cultures may express surprise emotion with different parts of facial muscles, again, challenging the hypothesis of emotion universality.

\section{Discussion}
While the universality of emotions may hold for certain facial movements, such as the AU12 (Lip Corner Puller), which shows consistency among both Asians and non-Asians—there are notable differences in the intensity of these movements and the specific regions activated during emotional expression. These variations may serve as early indicators of cultural or geographical influences on how individuals express emotions. Although some ambiguity may arise due to the simplifications made in this study, it nonetheless provides a foundational step toward a richer understanding of cross-cultural emotional expression. Future research could benefit from incorporating more ethnically annotated micro-expression datasets to further examine and potentially challenge the theory of emotional universality.

\section{Conclusion}
In this research, we have validated the influence of ethnicity context in expression analysis via a prima facie analysis. On top of that, we have constructed an ethnicity, gender and age-annotated CASME2-SAMM joint database, which can be further expanded in the future to include more micro-expression datasets. In a preliminary feature study, we have proposed a convolutional-based backbone to study the motion features and a patch-based backbone to study the face texture, which correlates to the ethnic context. The work can be further advanced via a causality study to identify the strength of ethnic context in influencing the expression analysis, and other contexts such as age and gender can be included as well in the next iteration of such work.

\section*{Acknowledgements}
This work was supported by the Research Council of Finland (former Academy of Finland) Academy Professor project EmotionAI (grants 336116, 359854), the University of Oulu \& Research Council of Finland Profi 7 (grant 352788), EU HORIZON-MSCA-SE-2022 project ACMod (grant 101130271), and Infotech Oulu. The authors wish to acknowledge CSC – IT Center for Science, Finland, for computational resources.

\bibliographystyle{IEEEbib}
\bibliography{main}

\end{document}